\documentclass[conference]{IEEEtran}
\IEEEoverridecommandlockouts

\usepackage[noadjust]{cite}  
\usepackage{amsmath,amssymb,amsfonts}
\usepackage{algorithmic}
\usepackage{graphicx}
\usepackage{textcomp}
\usepackage{xcolor}
\usepackage{tikz}

\def\BibTeX{{\rm B\kern-.05em{\sc i\kern-.025em b}\kern-.08em
    T\kern-.1667em\lower.7ex\hbox{E}\kern-.125emX}}
\DeclareMathOperator*{\argmax}{arg\,max}
\DeclareMathOperator*{\argmin}{arg\,min}
\newcommand{\todo}[1]{{\bfseries{\scshape{\color{orange}TODO: #1}}}}

\usepackage[subpreambles=true]{standalone}
\usepackage{cleveref}
\usepackage[ruled,vlined]{algorithm2e}
\usepackage{subfig}

\newcommand\copyrighttext{%
  \footnotesize \copyright 2021 IEEE.  Personal use of this material is permitted.  Permission from IEEE must be obtained for all other uses, in any current or future media, including reprinting/republishing this material for advertising or promotional purposes, creating new collective works, for resale or redistribution to servers or lists, or reuse of any copyrighted component of this work in other works.}
\newcommand\copyrightnotice{%
\begin{tikzpicture}[remember picture,overlay]
\node[anchor=south,yshift=10pt] at (current page.south) {\fbox{\parbox{\dimexpr\textwidth-\fboxsep-\fboxrule\relax}{\copyrighttext}}};
\end{tikzpicture}%
}

\begin{document}

\title{\raisebox{4mm}{\strut}Learning to Robustly Negotiate Bi-Directional Lane Usage in High-Conflict Driving Scenarios \\

\thanks{\IEEEauthorrefmark{1}Correspondence to christoph.killing(at)tum.de. This work was supported by the German Academic Exchange Service and the CMU Argo AI Center for Autonomous Vehicle Research} 

\author{\IEEEauthorblockN{Christoph Killing\IEEEauthorrefmark{1}\IEEEauthorrefmark{2},
Adam Villaflor\IEEEauthorrefmark{2} and John M. Dolan\IEEEauthorrefmark{2}}
\IEEEauthorblockA{\IEEEauthorrefmark{1}Technical University of Munich}
\IEEEauthorblockA{\IEEEauthorrefmark{2}The Robotics Institute, Carnegie Mellon University}
}
}


\maketitle
\copyrightnotice
\begin{abstract}
Recently, autonomous driving has made substantial progress in addressing the most common traffic scenarios like intersection navigation and lane changing. However, most of these successes have been limited to scenarios with well-defined traffic rules and require minimal negotiation with other vehicles.
In this paper, we introduce a previously unconsidered, yet everyday, high-conflict driving scenario requiring negotiations between agents of equal rights and priorities.
There exists no centralized control structure and we do not allow communications.
Therefore, it is unknown if other drivers are willing to cooperate, and if so to what extent. 
We train policies to robustly negotiate with opposing vehicles of an unobservable degree of cooperativeness using multi-agent reinforcement learning (MARL). 
We propose Discrete Asymmetric Soft Actor-Critic (DASAC), a maximum-entropy off-policy MARL algorithm allowing for centralized training with decentralized execution. 
We show that using DASAC we are able to successfully negotiate and traverse the scenario considered over $99\%$ of the time.
Our agents are robust to an unknown timing of opponent decisions, an unobservable degree of cooperativeness of the opposing vehicle, and previously unencountered policies. 
Furthermore, they learn to exhibit human-like behaviors such as defensive driving, anticipating solution options and interpreting the behavior of other agents.
\end{abstract}

\section{Introduction}
Recently, research on automated driving has focused on scenarios with a clearly defined set of traffic rules to which all vehicles are subject, such as intersection handling \cite{carolyn-hierarchy} and lane changing \cite{8968565}.
Yet, there also exist situations in which there are no well-defined traffic rules available. 
On residential roads, for instance, there might not be a dedicated lane per driving direction or sufficient road width for two opposing vehicles to pass by each other simultaneously. 
This leads to a situation in which autonomous agents have to negotiate subsequent bidirectional lane usage in what we call a high-conflict scenario. 

\subsection{High-Conflict Scenarios}
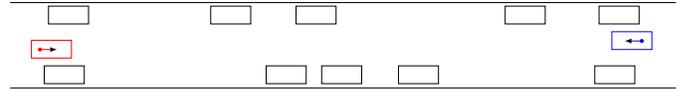
\begin{figure}[t]
    \centering
        \begin{tikzpicture}
    
    \draw [line width=0.1mm,  black] (0,0) -- (15.6,0);
    \draw [line width=0.1mm,  black] (0,2) -- (15.6,2);

    \draw[draw=black] (0.8,0.1) rectangle ++(0.938,0.417);  
    \draw[draw=black] (6.0,0.1) rectangle ++(0.938,0.417);  
    \draw[draw=black] (7.3,0.1) rectangle ++(0.938,0.417);  
    \draw[draw=black] (9.1,0.1) rectangle ++(0.938,0.417);  
    \draw[draw=black] (13.7,0.1) rectangle ++(0.938,0.417);  

    \draw[draw=black] (0.9,1.5) rectangle ++(0.938,0.417);  
    \draw[draw=black] (4.7,1.5) rectangle ++(0.938,0.417);  
    \draw[draw=black] (6.7,1.5) rectangle ++(0.938,0.417);  
    \draw[draw=black] (11.6,1.5) rectangle ++(0.938,0.417);  
    \draw[draw=black] (13.8,1.5) rectangle ++(0.938,0.417);  

    \draw[draw=red] (0.5,0.7) rectangle ++(0.938,0.417);  
    \fill[draw=red, fill=red] (0.7,0.9) circle [radius=1pt];
    \draw[>=latex,draw=red, ->] [line width=0.2mm] (0.7, 0.9) -- (1.1,0.9);

    \draw[draw=blue] (14.1,0.9) rectangle ++(0.938,0.417);  
    \fill[draw=blue, fill=blue] (14.8,1.1) circle [radius=1pt];
    \draw[>=latex,draw=blue, ->] [line width=0.2mm] (14.8, 1.1) -- (14.4,1.1);

    \end{tikzpicture}  
    \caption{Top-down view of proposed scenario: Neighborhood road on which two moving vehicles (red and blue) have to negotiate conflicting interests in a shared lane. 
    }
    \label{fig:intro_scenario}
\end{figure}

We consider a scenario simulating everyday neighborhood driving, as shown in Fig.~\ref{fig:intro_scenario}.
Two vehicles travel down the same lane but in opposing directions, with gaps to pull into between cars parked at either curb.
These cars locally reduce the width of the available road. 
It is up to the agents to negotiate the use of the available space in order to maneuver around each other without crashing into the parked vehicles.
{}
One can think of the resulting interactions as a game, even though both cars are constantly moving. 
The drivers are observing each other as well as their environment to make driving decisions. 
After one driver makes a decision, the other one is able to observe the trajectory of the first vehicle and react accordingly, which in turn influences the first driver's next decision, and so on.
We model the proposed scenario as a decentralized partially observable Markov Decision Process (dec-POMDP) \cite{oliehoek_amato_2016} where agents can only react to their local egocentric observations. 
There is no communication channel between them and no centralized control structure. 
Consequently, if one agent selects an action, there is no guarantee the other agent will select a compatible one. 
As in real-world driving, the control strategy as well as the degree of collaboration of the opposing vehicle is unknown. 

\subsection{Main Contributions}
In this paper, we demonstrate how to enable agents to robustly negotiate in the presented high-conflict driving scenario.
We model the interactions in a game-theoretic framework of discrete behavior decisions, which conventional control laws translate into throttle and steering commands. 
To facilitate the emergence of complex negotiation dynamics, we use an off-policy multi-agent reinforcement learning (MARL) approach. 
Our main contributions are:
(1) \textit{Scenario}: We examine an everyday and to the best of our knowledge so far unaddressed scenario challenging the social component of automated cars.
(2) \textit{Baselines}: 
As no common baselines exist, we devise two approaches to solve the scenario using rule-based methods. 
(3) \textit{Incentivizing Interaction}: We introduce a reward function parametrized by a cooperativeness parameter which continuously incentivizes agents to interact. 
We further demonstrate that by tuning this parameter, we achieve a corresponding and interpretable change in an agent's cooperativeness at the behavioral level. 
(4) \textit{Algorithm}: We propose Discrete Asymmetric Soft Actor-Critic (DASAC). DASAC leverages asymmetric information in a maximum-entropy off-policy actor-critic algorithm. It allows for a centralized learning process while enabling decentralized execution in a general-sum game.
We show that agents trained using our DASAC algorithm are highly robust to an unobservable degree of cooperativeness of the opposing vehicle.

\section{Preliminaries}

\subsection{Markov Games}
A Markov Game  \cite{webb07} with $K$ agents and a finite set of states $S$ is defined as $< K, S, A_1, ..., A_k, T, r_1, ..., r_k >$, including action set $A_k$ available to each agent $k$ ($k \in [1, ..., K]$), transition function T and reward function $r_k$ for each agent. 

There are two predominant action-selection scheduling mechanisms: In static games, each player selects its action, then a step is taken in the environment. In dynamic games, agents select their actions at different time steps. 
One can think of it as a turn-based game, but in general we can employ any underlying time-schedule to determine when a player takes its turn.
In this paper, we examine discrete behavior actions. 
These do not require a new behavior decision being made at every time step. 
Instead, we use conventional low-level controls to follow a selected behavior for multiple time steps.

A Markov game can have several reward-distribution mechanisms. 
When all agents receive the same reward $r_1 = ... = r_k = R$, a Markov game is called fully cooperative. 
In zero-sum games, one agent's win is the other agent's loss, such that the sum of rewards is zero. 
This is considered a competitive game. 
Both of these reward mechanisms introduce some notion of symmetry, providing an agent with additional information about other policies in the scenario. 
In a general-sum game, each agent attempts to maximize its own, independent reward.
As other agents in the scenario do not know about the behaviors incentivized, they cannot expect certain actions based on their own reward.

\subsection{Maximum-Entropy Reinforcement Learning}
In single-agent reinforcement learning (RL), we want to learn a policy $\pi(a_t|s_t)$ which maximizes the cumulative discounted reward at each time-step $t$ ($a_t \in A, s_t \in S$). 
Maximum entropy RL augments the reward with an entropy term \cite{10.5555/3305381.3305521}. The optimal policy aims to maximize a trade-off between its entropy and the reward at each visited state:

\begin{align}
    \pi^*_{\text{MaxEnt}} = \argmax_\pi \sum_t \mathbb{E}_{(s_t, a_t)\sim \rho_\pi}[r_t + \alpha \mathbb{H}(\pi(\cdot|s_t))]
\end{align}

Temperature $\alpha$ determines the relative importance of entropy and reward. 
The soft Bellman equation with discount factor $\gamma$ is consequently defined as:
\begin{align}
Q^\pi_{}(s_t, a_t) = r_t + \gamma (&\mathbb{E}_{\pi}\left[Q^\pi_{}(s_{t+1}, a_{t+1})\right] \nonumber\\ &+ \alpha \cdot \mathbb{H}(\pi_{}(\cdot|s_{t+1})))
\label{eq:bellman_soft_policyiteration}
\end{align}

The Boltzmann policy $\pi^B \propto exp(Q/\alpha)$ is shown to be the optimal policy satisfying \eqref{eq:bellman_soft_policyiteration} \cite{10.5555/3305381.3305521}. 

\section{Related Work}
\label{sec:relatedwork}
The introduced problem of learning discrete high-level behavior strategies for automated driving in multi-agent systems is twofold. 
Firstly, we give a general overview of RL methods.
Secondly, we present applications of single- and multi-agent learning to autonomous driving scenarios. 

\subsection{Reinforcement Learning}
\label{section:realted_rl}
Within the class of discrete RL algorithms, Q-learning is a popular choice for its robustness. 
While the original approaches date back several decades, only recently Mnih et al. \cite{mnih2015humanlevel} were able to show that Q-learning can achieve super-human performance in playing Atari games when using a Neural Network to approximate the Q-function. 
Yet, many real-world problems cannot be sufficiently modeled by single-agent RL, as there are multiple actors. 
One approach to generate actions for several agents is to train one policy that perceives all observations and produces them jointly, generating highly coordinated behavior. 
However, two problems arise in this centralized setup: 
Firstly, it implies an immense amount of communication, making the approach infeasible for real-world automated driving. 
Secondly, it is not scalable to a large number of agents, as the dimension of the joint action-space increases exponentially with their number.
Another approach is decentralized execution, where each agent follows its own independent policy. 
However, the action of a single agent can lead to different outcomes throughout the training process, making the transition of the environment non-stationary from the perspective of each agent.
This so-called dec-POMDP problem is intractable \cite{10.5555/3237383.3238080}, as there are no guarantees that all agents select compatible actions.

Several approaches for decentrally executed MARL in discrete action spaces exist. 
The naive approach is to train independent policies for each agent \cite{DBLP:conf/icml/Tan93, 10.1371/journal.pone.0172395}. 
Yet, this independent Q-learning approach is prone to instabilities originating in the non-stationarity of the environment. 
Fingerprinting samples in the replay buffer is shown to improve convergence \cite{10.5555/3305381.3305500}. 
We use this as a baseline to compare to later on. 
Centralized training with decentralized execution can improve convergence by providing a centralized critic with state information at training time, but allowing decentralized execution purely based on an agent's local observations at test time. 
Existing discrete-action centralized learning approaches which allow decentralized execution are deterministic and place symmetry requirements on the reward function through assuming fully cooperative or fully competitive games \cite{10.5555/3237383.3238080, Rashid2018QMIXMV, Foerster2018CounterfactualMP}.
There also exist various centralized training approaches in continuous action spaces, which can be executed decentrally \cite{DBLP:conf/nips/LoweWTHAM17, Li2019RobustMR, Yang2020CM3, multiagentsoft}. 
These use policy iteration \cite{10.5555/551283} for centralized training with decentralized execution by providing state information to a critic, but only local observations to an actor. 

We bring the stability of centralized training with decentralized execution in a discrete general-sum game to a maximum-entropy framework.
Such non-deterministic policies benefit from improved exploration, as actions are sampled from them rather than greedily selected, and are able to keep track of multiple solution modes.
This promises to be beneficial in our reward setting. 
We base our derivation of DASAC on Soft Q-learning (SQL) \cite{10.5555/3305381.3305521}, a stochastic variant of discrete deep Q-learning. 
SQL can be transformed to continuous action spaces using Stein Variational Gradient Descent \cite{10.5555/3305381.3305521}. 
Subsequently, \cite{Haarnoja2018SoftAO} proposed Soft Actor-Critic (SAC) to improve the performance. 
SAC can be adapted to work in discrete action spaces, as \cite{christodoulou2019soft} showed for Atari.
We found \cite{christodoulou2019soft}'s algorithm to be the single-agent symmetric version of our approach. 
Yet, as \cite{christodoulou2019soft} only compares performance to DQN and not SQL, the benefits of formulating SQL as an actor-critic algorithm for Atari environments remain unclear. 
The potential of our proposed algorithm is twofold: 
Firstly, it lies in the possibility of providing asymmetric observations to the actor and the critic. 
Secondly, due to the delay induced by policy improvement \cite{10.5555/551283}, the actions of other agents in a multi-agent setup change more slowly. 
Consequently, agents can learn to anticipate other agent's behaviors from their observations, even though the underlying reward is unknown. 

\subsection{Learning for Automated Driving} 
\label{sec:related:ad}
Learning approaches are increasingly being researched in the field of automated driving since they generalize well to previously unencountered situations. 
Plenty of recent work addresses single-agent RL problems for automated driving, such as intersections \cite{carolyn-hierarchy} and lane changes \cite{8968565}. 
Significantly less work applies multi-agent learning to autonomous driving.

The issue of unknown driver behaviors in a MARL setting is addressed by \cite{DBLP:journals/corr/abs-1903-01365} and \cite{Hu2019InteractionawareDM}, who define driver type variables influencing aggressiveness of longitudinal control through a parametrized reward function. 
While they are able to show that traversal time decreases for a more aggressive ego-agent, they do not evaluate robustness towards different aggressiveness settings of the other vehicles. 
Our parametrized reward function incentivizes continuous interactions; they only focus on the behavior of the ego-agent \cite{DBLP:journals/corr/abs-1903-01365} or only take other vehicles into account when safety-relevant actions, such as hard braking, were induced in them by the ego-agent \cite{Hu2019InteractionawareDM}.

The underlying game-theoretic interactions of an on-ramp merger are examined by \cite{8916951} in a \textit{T}-shaped grid-world environment.
They evaluate turn-taking agents compared to simultaneously playing ones. 
While their approach of considering interactions is similar to ours, their turn-taking evaluation has severely limited applicability to real-world driving, as one vehicle waits for the other to take its turn. 
Furthermore, their merging vehicle has additional information on the vehicle in traffic. 
We take their game-theoretic approach a step further and randomize the action-schedule, representing the unknown timing of decisions in real-world driving, while allowing both vehicles to continuously move.

In contrast to the multi-agent works mentioned here, our work is geared more towards the high-level reasoning and interactions than learning low-level control strategies.
We furthermore also learn lateral control, such as \cite{Tang2019TowardsLM}, which uses curriculum learning to address a MARL highway merger. 

\section{Approach}
\subsection{Problem Formulation}
\label{sec:problem_form}
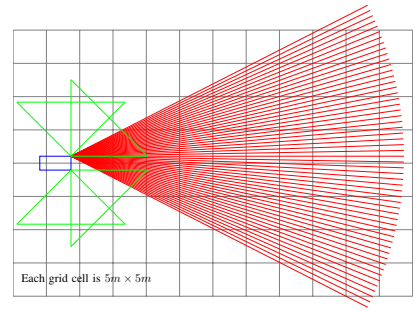
\begin{figure}[h]
    \centering
        \begin{tikzpicture}
    \draw[step=1cm,gray,very thin] (-1, -4) grid (11,4);
    
    \node[] at (1.2, -3.5) {Each grid cell is $5m \times 5m$};
    
    \draw[draw=blue] (-0.2,-0.417/2) rectangle ++(0.938,0.417);  

    \coordinate (radar_origin) at (0.938-0.2,0.417/2);
    \draw [red] (radar_origin) -- +(27:10cm);
    \draw [red] (radar_origin) -- +(-27:10cm);
    
    \draw [red] (radar_origin) -- +(-26:10cm);
    \draw [red] (radar_origin) -- +(-25:10cm);
    \draw [red] (radar_origin) -- +(-24:10cm);
    \draw [red] (radar_origin) -- +(-23:10cm);
    \draw [red] (radar_origin) -- +(-22:10cm);
    \draw [red] (radar_origin) -- +(-21:10cm);
    \draw [red] (radar_origin) -- +(-20:10cm);
    \draw [red] (radar_origin) -- +(-19:10cm);
    \draw [red] (radar_origin) -- +(-18:10cm);
    \draw [red] (radar_origin) -- +(-17:10cm);
    \draw [red] (radar_origin) -- +(-16:10cm);
    \draw [red] (radar_origin) -- +(-15:10cm);
    \draw [red] (radar_origin) -- +(-14:10cm);
    \draw [red] (radar_origin) -- +(-13:10cm);
    \draw [red] (radar_origin) -- +(-12:10cm);
    \draw [red] (radar_origin) -- +(-11:10cm);
    \draw [red] (radar_origin) -- +(-10:10cm);
    \draw [red] (radar_origin) -- +(-9:10cm);
    \draw [red] (radar_origin) -- +(-8:10cm);
    \draw [red] (radar_origin) -- +(-7:10cm);
    \draw [red] (radar_origin) -- +(-6:10cm);
    \draw [red] (radar_origin) -- +(-5:10cm);
    \draw [red] (radar_origin) -- +(-4:10cm);
    \draw [red] (radar_origin) -- +(-3:10cm);
    \draw [red] (radar_origin) -- +(-2:10cm);
    \draw [red] (radar_origin) -- +(-1:10cm);

    \draw [red] (radar_origin) -- +(26:10cm);
    \draw [red] (radar_origin) -- +(25:10cm);
    \draw [red] (radar_origin) -- +(24:10cm);
    \draw [red] (radar_origin) -- +(23:10cm);
    \draw [red] (radar_origin) -- +(22:10cm);
    \draw [red] (radar_origin) -- +(21:10cm);
    \draw [red] (radar_origin) -- +(20:10cm);
    \draw [red] (radar_origin) -- +(19:10cm);
    \draw [red] (radar_origin) -- +(18:10cm);
    \draw [red] (radar_origin) -- +(17:10cm);
    \draw [red] (radar_origin) -- +(16:10cm);
    \draw [red] (radar_origin) -- +(15:10cm);
    \draw [red] (radar_origin) -- +(14:10cm);
    \draw [red] (radar_origin) -- +(13:10cm);
    \draw [red] (radar_origin) -- +(12:10cm);
    \draw [red] (radar_origin) -- +(11:10cm);
    \draw [red] (radar_origin) -- +(10:10cm);
    \draw [red] (radar_origin) -- +(9:10cm);
    \draw [red] (radar_origin) -- +(8:10cm);
    \draw [red] (radar_origin) -- +(7:10cm);
    \draw [red] (radar_origin) -- +(6:10cm);
    \draw [red] (radar_origin) -- +(5:10cm);
    \draw [red] (radar_origin) -- +(4:10cm);
    \draw [red] (radar_origin) -- +(3:10cm);
    \draw [red] (radar_origin) -- +(2:10cm);
    \draw [red] (radar_origin) -- +(1:10cm);
    \draw [red] (radar_origin) -- +(0:10cm);

    \draw [green] (radar_origin) -- +(90:2.3cm) coordinate (A);
    \draw [green] (radar_origin) -- +(0:2.3cm) coordinate (B);
    \draw [green] (A) -- (B);
    
    \draw [green] (radar_origin) -- +(45:2.3cm) coordinate (C);
    \draw [green] (radar_origin) -- +(135:2.3cm) coordinate (D);
    \draw [green] (C) -- (D);

    \coordinate (ultra_origin_right) at (0.938-0.2,-0.417/2);
    \draw [green] (ultra_origin_right) -- +(-90:2.3cm) coordinate (E);
    \draw [green] (ultra_origin_right) -- +(0:2.3cm) coordinate (F);
    \draw [green] (E) -- (F);

    \draw [green] (ultra_origin_right) -- +(-45:2.3cm) coordinate (G);
    \draw [green] (ultra_origin_right) -- +(-135:2.3cm) coordinate (H);
    \draw [green] (G) -- (H);
    \end{tikzpicture}        
    \caption{Observation space of an agent. Visualized are the fields of view (fov) of ultrasonic sensors (green; returning distance of closest object in fov) and a radar (red; visualization scaled by 0.5; returning distance and relative velocity per ray).} 
    \label{fig:obs}
\end{figure}

\begin{figure}[h]
    \centering
        \begin{tikzpicture}
    \draw (-2,5) -- (2,5) -- (2,4) -- (-2,4) -- (-2,5);
    \node[] at (0,4.5) {\footnotesize Strategic Level};
    
    \node[] at (0,3.5) {\scriptsize Route Speed Criteria};
    \draw[->] (-1.5,4) -- (-1.5,3);
    \draw[->] (1.5,4) -- (1.5,3);
    
    \draw (-2,2) -- (2,2) -- (2,3) -- (-2,3) -- (-2,2);
    \node[] at (0,2.5) {\footnotesize Behavior Level};
    
    \node[] at (0,1.5) {\scriptsize Feedback Criteria};
    \draw[->] (-1.5,2) -- (-1.5,1);
    \draw[->] (1.5,2) -- (1.5,1);

    \draw (-2,0) -- (2,0) -- (2,1) -- (-2,1) -- (-2,0);
    \node[] at (0,0.5) {\footnotesize Control Level};
    
    \node[text width=2cm] at (-4.5,2.7) {\footnotesize Environmental Input};
    \coordinate (env_int_low) at (-3.5,2.5);
    \coordinate (con_in_low) at (-2,2.5);
    \draw[->] (env_int_low) -- (con_in_low);

    \node[text width=2cm] at (-4.5,0.4) {\footnotesize Environmental Input};
    \coordinate (env_int_low) at (-3.5,0.2);
    \coordinate (con_in_low) at (-2,0.2);
    \draw[->] (env_int_low) -- (con_in_low);
    
    \coordinate (strat_in_low) at (-2,4.25);
    \coordinate (r1_uppercorner) at (-3.75,3.75);
    \coordinate (r1_lowercorner) at (-3.75,0.4);
    \coordinate (r1_env_out) at (-2,0.4);
    \coordinate (strat_in_con_low) at (-3,4.25);
    
    \coordinate (o2in) at (-2,2.75);
    \coordinate (man_out_high) at (-3,2.75);
    \coordinate (o2_lowercorner) at (-3.5,0.6);
    \coordinate (o2_env_out) at (-2,0.6);
    \coordinate (o2_env_out_connection) at (-3,0.6);
    
    \coordinate (r2in) at (-2,2.25);
    \coordinate (r2_lowercorner) at (-3,0.75);
    \coordinate (r2_env_out_connection) at (-2.75,0.8);

    \coordinate (man_out_low) at (-3,2.25);
    \coordinate (con_out_high) at (-2,0.75);

    \draw[->] (r2_lowercorner) -- (con_out_high);
    \draw[-] (r2_lowercorner) -- (man_out_low);
    \draw[-] (man_out_low) -- (r2in);

    \draw[-] (man_out_high) -- (o2in);
    \draw[-] (man_out_high) -- (strat_in_con_low);
    \draw[->] (strat_in_con_low) -- (strat_in_low);

    \draw[->] (2,4.5) -- (3,4.5);
    \node[text width=3cm] at (4.7,4.5) {\footnotesize General Plans};
    \node[text width=3cm] at (8,4.5) {\footnotesize Traversing scenario at cooperativeness c}; 

    \draw[->] (2,2.75) -- (3,2.75);
    \draw[-] (2.5,2.75) -- (2.5,2.25);
    \draw[<-] (2,2.25) -- (2.5,2.25);
    \node[text width=3cm] at (4.7,2.5) {\footnotesize Controlled Action Patterns};
    \node[text width=3cm] at (8,2.5) {\footnotesize $\mathbb{E}[f_{\text{behavior}}]=4$Hz}; 
    
    \draw[->] (2,0.75) -- (3,0.75);
    \draw[-] (2.5,0.75) -- (2.5,0.25);
    \draw[<-] (2,0.25) -- (2.5,0.25);
    \node[text width=3cm] at (4.7,0.5) {\footnotesize Automatic Action Patterns};
    \node[text width=3cm] at (8,0.5) {\footnotesize $f_{\text{controls}} = 20$Hz};

    \end{tikzpicture}  
    \caption{Hierarchical structure of the road user task (after \cite{michon1985critical})}
    \label{fig:road_user}
\end{figure}
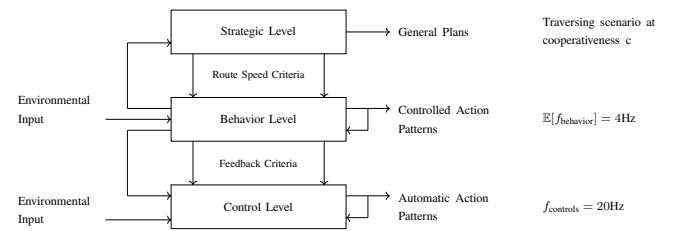

The considered dec-POMDP high-conflict scenario for automated driving consists of a straight road, on which vehicles are parked at random positions at either curb. 
Two agents driving in opposing directions approach each other. 
Contrary to the previously discussed well-researched scenarios,
the minimum set of environment features relevant to solving our scenario is unknown.
Therefore, each agent perceives a high-dimensional egocentric sensor-realistic observation, as shown in Fig.~\ref{fig:obs}, augmented by its cooperativeness, lateral and longitudinal position with respect to its driving direction, and steering and acceleration values. 
We allow agents to laterally position themselves in the \textit{shared lane} ($y_{\text{shared}}=4.5m$ {from respective right curb}), in which the opposing vehicle is approaching, or pull over to the right into their respective \textit{ego lane} ($y_{\text{ego}}=2.1m$ {from respective right curb}). 
No agent can pull into the lane to its far left, referred to as the \textit{opposing lane}, making right-hand driving rules inherent to the problem statement.
We model the goals and timing of decisions in our approach based on the hierarchical structure of the road user task shown in Fig.~\ref{fig:road_user}.
It outlines how high-frequency controls are conditioned on lower-frequency behavior decisions. 
In real-world driving, it is unknown to each driver \textit{at the time of their decision}, \textit{if} the opposing driver also made a new decision or \textit{when} the opposing driver's last decision was. 
We transfer this to the model of our problem by, after each behavior generation, sampling the number of time-steps until the next decision from a uniform distribution $\mathbb{U} \sim [4,5,6]$ in a $20$Hz simulation.

\subsection{Baselines}
To the best of our knowledge, no rule-based decision makers exist for traversing our scenario. To allow comparison with our learning approach, we propose two baselines:
\subsubsection{Threshold-Based}
This driving policy pulls into the next available gap once the space to the right is free and the opposing vehicle is closer than some distance threshold. 
\subsubsection{Reachability-Based}
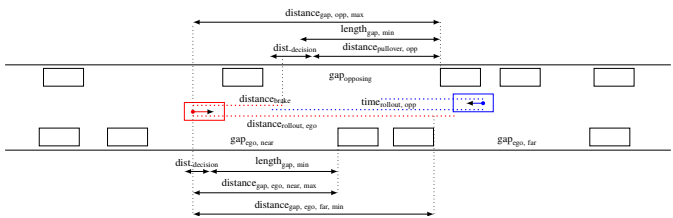
\begin{figure}[b]
    \centering
        \begin{tikzpicture}

    \draw [line width=0.1mm,  black] (0,0) -- (15.6,0);
    \draw [line width=0.1mm,  black] (0,2) -- (15.6,2);

    \draw[draw=black] (0.8,0.1) rectangle ++(0.938,0.417);  
    \draw[draw=black] (2.4,0.1) rectangle ++(0.938,0.417);  
    \draw[draw=black] (7.8,0.1) rectangle ++(0.938,0.417);  
    \draw[draw=black] (9.1,0.1) rectangle ++(0.938,0.417);  
    \draw[draw=black] (13.7,0.1) rectangle ++(0.938,0.417);  

    \draw[draw=black] (0.9,1.5) rectangle ++(0.938,0.417);  
    \draw[draw=black] (5.1,1.5) rectangle ++(0.938,0.417);  
    \draw[draw=black] (10.2,1.5) rectangle ++(0.938,0.417);  
    \draw[draw=black] (11.6,1.5) rectangle ++(0.938,0.417);  
    \draw[draw=black] (13.8,1.5) rectangle ++(0.938,0.417);  

    \draw[draw=red] (4.2,0.7) rectangle ++(0.938,0.417);  
    \coordinate (red_origin) at (4.4, 0.9);
    \fill[draw=red, fill=red] (red_origin) circle [radius=1pt];
    \draw[>=latex,draw=red, ->] [line width=0.2mm] (red_origin) -- (4.9,0.9);

    \draw[draw=blue] (10.5,0.9) rectangle ++(0.938,0.417);  
    \coordinate (blue_origin) at (11.2, 1.1);
    \fill[draw=blue, fill=blue] (blue_origin) circle [radius=1pt];
    \draw[>=latex,draw=blue, ->] [line width=0.2mm] (blue_origin) -- (10.8,1.1);

    \node[] at (5.8, 0.2) {\scriptsize $\text{gap}_{\text{ego, near}}$};
    \node[] at (12, 0.2) {\scriptsize $\text{gap}_{\text{ego, far}}$};

    \draw[dotted] (red_origin) -- (4.4, -1.5);
    \draw[dotted] (7.8, 0.1) -- (7.8, -1);
    \draw[dotted] (10.05, 0.8) -- (10.05, -1.5);

    \draw[thick, dotted, draw=red] (red_origin) ++(0, -0.1) -- (10.55, 0.8); 
        \node[] at (6.5,0.6) {\scriptsize $\text{distance}_{\text{rollout, ego}}$};
        
    \draw[thick, dotted, draw=blue] (blue_origin) ++ (0, -0.15)-- (6.2, 0.95); 
        \node[] at (9,1.1) {\scriptsize $\text{time}_{\text{rollout, opp}}$};

    \draw[>=latex,<->] [line width=0.2mm] (4.8, -0.5) -- (7.8,-0.5);  
            \node[] at (6.5,-0.3) {\scriptsize $\text{length}_{\text{gap, min}}$};
    \draw[>=latex,<->] [line width=0.2mm] (4.2, -0.5) -- (4.8,-0.5);  
            \node[] at (4.5,-0.3) {\scriptsize $\text{dist.}_{\text{decision}}$};

    \draw[>=latex, <->] [line width=0.2mm] (red_origin)++(0, -1.9) -- (7.8,-1); 
        \node[] at (6.2,-0.8) {\scriptsize $\text{distance}_{\text{gap, ego, near, max}}$};

    \draw[>=latex, <->] [line width=0.2mm] (red_origin)++(0, -2.4) -- (10.05,-1.5); 
        \node[] at (6.9,-1.3) {\scriptsize $\text{distance}_{\text{gap, ego, far, min}}$};

    \draw[dotted] (red_origin)++(0, 2) -- (red_origin);
    \draw[dotted] (6.5, 1.1) -- (6.5, 2.2);
    \draw[dotted] (10.2, 1.8) -- (10.2, 3);
    \node[] at (8.1,1.7) {\scriptsize $\text{gap}_{\text{opposing}}$};
    \draw[thick, dotted, draw=red] (red_origin) ++(0, 0.15) -- (6.5, 1.05); 
        \node[] at (6.1,1.2) {\scriptsize $\text{distance}_{\text{brake}}$};
        
    \draw[thick, dotted, draw=blue] (blue_origin)++(0, 0.1) -- (8.8, 1.2); 

    \draw[>=latex,<->] [line width=0.2mm] (7.2, 2.2) -- (10.2,2.2);  
            \node[] at (8.8,2.4) {\scriptsize $\text{distance}_{\text{pullover, opp}}$};
            
    \draw[>=latex,<->] [line width=0.2mm] (6.2, 2.2) -- (7.2,2.2);  
            \node[] at (6.8,2.4) {\scriptsize $\text{dist.}_{\text{decision}}$};

    \draw[>=latex, <->] [line width=0.2mm] (red_origin)++(0, 2.1) -- (10.2,3); 
        \node[] at (7.5,3.2) {\scriptsize $\text{distance}_{\text{gap, opp, max}}$};

    \draw[>=latex,<->] [line width=0.2mm] (red_origin)++(2.5, 1.7) -- (10.2,2.6);  
            \node[] at (8.5,2.8) {\scriptsize $\text{length}_{\text{gap, min}}$};

    \end{tikzpicture}
    \caption{Qualitative visualization of decision parameters for red vehicle as ego-agent. 
    It perceives an upcoming collision when heading for the further gap in its ego lane. 
    The remaining options are to either pull over or remain in the shared lane. 
    Here, the red vehicle maximizes its own progress by braking in the shared lane, allowing the opposing car to pull over. 
    }
    \label{fig:reach}
\end{figure}

We roll out a kinematic single-track model (i.e. \cite{commonroad}) of the ego-agent for a distance depending on the parked vehicles at either curb. 
For the opposing agent, we roll out the model for the same number of time-steps to identify possible points of collision.
Decision parameters are visualized qualitatively in Fig.~\ref{fig:reach}. 
This approach results in an agent continuously maximizing its progress towards the furthest reachable Nash-Equilibrium \cite{webb07, reachability}.

\subsection{Varying Driver Behaviors}
\label{sec:reward}
We use a parametrized reward function $r(c)$ to transfer a cooperativeness parameter $c$ into varying driver behaviors.
It continuously incentivizes agents to interact through taking both of their velocities into account, as shown in \eqref{reward}. 
Each agent is able to observe its own cooperativeness, but not that of the opposing vehicle.
Highly cooperative agents with $c=0.5$ are consequently indifferent to which agent makes progress first, and drive with more anticipation. 
Uncooperative agents with $c=0$ are only rewarded for their own progress. 
\begin{align}
    r(c) \propto (1-c)\cdot v_{ego} + c \cdot v_{opp} \ \ \ c \in [0, 0.5]
    \label{reward}
\end{align}

\subsection{Curriculum Learning}
\begin{figure}[t]
    \centering
    \subfloat[][Four options - focus on negotiations \label{fig:six_veh_init_sub_a}] 
    {\scalebox{0.192}{\input{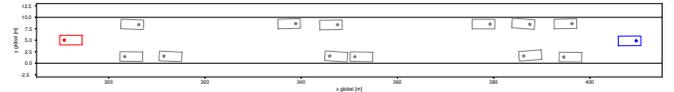}}}
    \hfill
    \subfloat[][Two options - focus on driving with anticipation \label{fig:six_veh_init_sub_b}] 
    {\scalebox{0.192}{\input{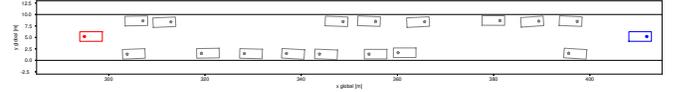}}}
    \caption{Example scenario initializations}
    \label{fig:six_veh_init}
\end{figure}

There exist two main challenges agents have to learn how to solve to successfully traverse the scenario presented:
Firstly, they need to learn to negotiate bidirectional lane usage. 
Secondly, they need to learn to anticipate the other driver's intentions. 
We decompose these into a curriculum learning approach. 
Intuitively, the more cars are parked at either curb, the fewer gaps allowing the vehicles to negotiate the use of the shared lane exist. 
Consequently, agents need to be more refined the more vehicles are parked on the road, as exploratory actions can easily prevent successful traversals. 
We show two example initializations in Fig.~\ref{fig:six_veh_init}. 
The probabilities of vehicles sampled are shown in Table~\ref{table:environment_init}. 
After learning basic driving capabilities in stage A, we iterate between stages B and C to prevent agents from associating a particular environment population with a behavior of the other agent, as that changes throughout the training process.

\begin{table}[t]
\centering
\caption[Curriculum Initialization]{Sampling probabilities of $p(no_{\text{vehicles}})$, the number of parked vehicles \textit{per side}.}
\begin{tabular}{l| c c c }
    & \multicolumn{3}{c}{$p(no_{\text{vehicles}})$} \\
    stage & 6 & 7 & 8 \\ \hline
    \textit{A} & 1 &  &  \\
    \textit{B} & 0.8 & 0.1 & 0.1 \\
    \textit{C} & 0.5 & 0.3 & 0.2  \\
\end{tabular}
\label{table:environment_init}
\end{table}

\subsection{Discrete Asymmetric Soft Actor-Critic}
The general idea of our DASAC is to reformulate SQL as an Actor-Critic algorithm where we explicitly represent the soft Q-function $Q$ and the policy $\pi$  with separate neural networks.
This allows us to provide them with different perceptions.
An acting policy $\pi$ should, at test time, only have access to its local observations and not the internal state of other agents.
However, knowledge of the internal state of other agents can reduce the effect of environment non-stationarity in MARL. 
We therefore train a state-based Q-network in a maximum-entropy framework and minimize the divergence between the actions of the acting policy $\pi$ and the Boltzmann policy.
This can also be thought of as behavioral cloning on local observations.

We augment the observation space discussed in Sec.~\ref{sec:problem_form} by the cooperativeness, steering angle, and acceleration of the opposing vehicle. We use this as the state-based perception of the $Q$-function. 
Through this knowledge of both vehicles' long-term goals and immediate commands, our critic is able to stabilize and coordinate the learning process. 
As the policy $\pi$ does not have access to this extra information, it learns to be robust towards varying behaviors of the opposing vehicle. 

Reformulating \eqref{eq:bellman_soft_policyiteration}, we can explicitly state this asymmetry of information through a policy depending on local observations $\pi(\cdot|o)$ and a Q-function with state information $Q(s, a)$. In policy iteration step $k$, we update $Q$ according to:

\begin{align}
    Q^\pi_{k}(s_t, a_t) &= r_t + \gamma \sum_{a'} \pi_{k-1}(a'| o_{t+1})\nonumber \\ 
    &\cdot \left[Q^\pi_{k-1}(s_{t+1}, a') - \alpha \cdot log(\pi_{k-1}(a'|o_{t+1}))\right] \label{eq:policy_evaluation_obs}
\end{align}

We can convert the learned Q-values into a probability distribution over actions using the Boltzmann policy $\pi_{k}^B \propto exp(Q^{\pi}_k/\alpha)$, which in the discrete case is a \textit{softmax} function. 
Using \eqref{eq:policy_evaluation_obs} as the current estimate of the true $Q$, $\pi_{k}^B$ is the policy which maximizes the soft Bellman equation in \eqref{eq:bellman_soft_policyiteration}. 
Then, we update our policy $\pi_k(\cdot|o)$ by minimizing its KL-divergence from $\pi_k^B(\cdot|s)$.

\begin{align}
    \pi_{k}(\cdot|o) = \argmin_{\pi} KL(\pi(\cdot|o) | \pi^B_{k}(\cdot|s)) \label{eq:policy_improvement}
\end{align}

If we run \eqref{eq:policy_improvement} to convergence in every policy iteration step, the proof of convergence of SQL \cite{10.5555/3305381.3305521} still holds. 
If we do not, we have a soft policy iteration just like SAC, which \cite{Haarnoja2018SoftAO} shows to converge. 
The update for one agent is given in Algorithm \ref{algoithm:dSAC}. 
In principle, this algorithm can be used whenever one might need to use different observations for a critic and an actor.

\section{Experiments}
\subsection{Setup}
\subsubsection{Action Space}
The action space consists of three behavior decisions, namely following the shared lane ($y_{\text{goal}}=y_{\text{shared}}$, $v_{\text{goal}}=8\frac{m}{s}$), pulling over ($y_{\text{goal}}=y_{\text{ego}}$, $v_{\text{goal}}=0$ if distance to next parked car in ego-lane $< 10m$ else $v_{\text{goal}}=2\frac{m}{s}$) and coming to a halt in the shared lane ($y_{\text{goal}}=y_{\text{shared}}$, $v_{\text{goal}}=0$).
We use conventional low-level controls \cite{4282788} to translate these set-points into steering and throttle commands.
\subsubsection{Reward Function} 
Once the opposing vehicle is closer than $d_{opp}$, we incentivize interactions with ego-cooperativeness $c$ using the reward function shown in \eqref{eq:rewardfunction}. 
A physically inspired collision penalty leads the agent to drive more slowly in critical situations. 

\begin{align}
    \label{eq:rewardfunction}
    r (t, c) = 
    \begin{cases}
        \frac{(1-c) \cdot v_{\text{ego}}(t) + c \cdot v_{\text{opp}}(t)}{10} & \text{if $d_{\text{opp}} < 80m$}\\
        +8 & \text{if success}\\
        -max(3, v_{\text{ego}}(t)) & \text{if collision}\\
        -3 & \text{if timeout}\\
        \frac{v_{\text{ego}}(t)}{10} & \text{else}\\
    \end{cases}
\end{align}

\subsection{Numerical Results}
\begin{figure}[t]
    \centering
    \scalebox{0.58}{\input{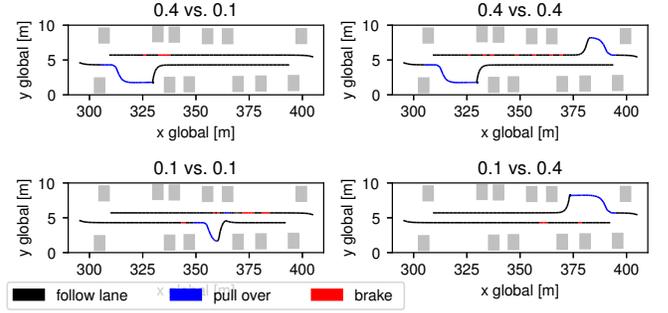}}
    \caption{Influence of cooperativeness on interaction $c_{\text{ego}}$ vs. $c_{\text{opp}}$. 
    Ego-agent driving from left to right. 
    Policy evaluated in self-play. 
    Behavior-level decisions visualized in path.}
    \label{fig:trajectoreis}
\end{figure}

We train policies in self-play and randomly sample the cooperativeness of each vehicle from a uniform distribution $\mathbb{U}\sim[0, 0.5]$. 
The observed cooperativeness influences agent behavior in a distinct and intuitively explainable way, as visualized in Fig.~\ref{fig:trajectoreis}. 
Highly cooperative agents drive with more anticipation than less cooperative ones, which instead attempt to maximize their own progress.
Traversal times consequently increase with increasing cooperativeness. 

\begin{figure}[h]
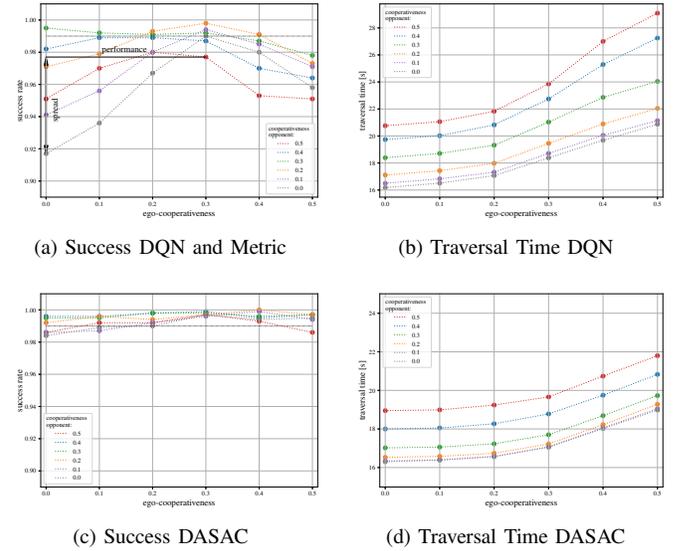

    \centering
    \subfloat[][Success DQN and Metric \label{fig:metric}]{\scalebox{0.24}{\input{figures/metric.pgf}}}
    \hfill
    \subfloat[][Traversal Time DQN \label{fig:dqn_time}]{\scalebox{0.24}{\input{figures/cooperation_plot_time.pgf}}}\\
    \subfloat[][Success DASAC \label{fig:centralized_coop_results_sub_a}]{\scalebox{0.24}{\input{figures/DASAC_30_success_medium.pgf}}}
    \hfill
    \subfloat[][Traversal Time DASAC \label{fig:centralized_coop_results_sub_b}]{\scalebox{0.24}{\input{figures/DASAC_30_time_medium.pgf}}}
    \caption{Self-play evaluations of peak-performing policies. We show success-rate and traversal time over cooperativeness of the ego-agent for various cooperativeness values of the opposing agent. 
    Each data point results from a cooperativeness pairing simulated on a test set of $1000$ stage $B$ initializations.
    }
    \label{fig:evaluation}
\end{figure}

We aim to identify an ego-cooperativeness value that is highly robust towards any cooperativeness of the opposing vehicle.
For a trained policy, we evaluate six values of cooperativeness per vehicle, resulting in 36 cooperativeness pairings. 
Two full self-play evaluations of peak-performing policies are shown in Fig.~\ref{fig:evaluation}. 
We find that our DASAC algorithm is significantly more robust towards an unobservable degree of cooperativeness of the opposing vehicle than a fingerprinted DQN policy.
It arrives at a success rate of $99.5\%$ at $c_{ego}=0.3$ given any cooperativeness of the opposing vehicle and also completes traversals faster. 
We attribute both to the stabilized learning through centralized critics. 

We compare various approaches by performing this analysis for every $100^{th}$ policy in between training epoch $1500$ and $2500$ from three seeds per algorithm. Mean results are shown in Table~\ref{tab:mean_performance_medium_test}. 
A training epoch consists of 32 environments simulated in parallel and 2000 gradient steps for DQN, SQL and DASAC-$Q$ (two $\pi$ updates per $Q$ update).
Besides DQN, we also evaluate a decentrally trained SQL to quantify the effect of maximum-entropy learning. 
We find that SQL performs comparably to the DQN, but with greater robustness. 
We assume this originates from the ability to keep track of multiple solution modes and the improved exploration. 
We furthermore observe that our DASAC quite significantly outperforms all other considered approaches. 
It achieves the highest success at a high robustness towards extreme pairings of cooperativeness. 
We attribute this to the centralized training and the maximum-entropy learning. 
We are also able to confirm the favorable influence of our curriculum learning approach. 
The accompanying increase in spread, we speculate, is due to the agents lacking the ability to drive with sufficient anticipation. 

\begin{table}[t]
    \centering
    \caption{Relative performance of algorithms and spread towards extreme cooperativeness pairings. All baselines in self-play. 
    Performance and spread are reported as shown in Fig.~\ref{fig:metric}. 
    We use dueling network architectures, target networks, and prioritized experience replay for all algorithms.
    }
    \begin{tabular}{l|rr}
        &  performance [$\%$]& spread [$\%$] \\ \hline
        Threshold 90m & 89.10 &-\\
        Reachability & 96.40 &-\\ \hline \hline
        DQN & 97.30 & 6.31\\
        SQL & 97.11 & 5.74\\ \hline
        DASAC (no curriculum)& 98.36 & 6.45\\
        \textbf{DASAC} (self-play) & \textbf{99.09} & 5.53\\
        DASAC (unencountered policy) & {99.05} & 3.06\\
    \end{tabular}
    \label{tab:mean_performance_medium_test}
\end{table}

When confronting previously unencountered policies from different training seeds, we see no major negative impact on performance. 
In fact, due to the massive reduction in the spread of results, we find that unencountered policies are more robust than policies evaluated in self-play. 
We cannot blame individual policies for unsuccessful traversals, as multi-agent credit assignment is an open problem of MARL \cite{DBLP:journals/corr/abs-1906-04737}. 
However, we assume the improvement originates from self-play situations in which a shortcoming in the policy reduces performance of both agents.
For unencountered policies, one policy having a learning deficit seems to be compensated for by the other one, which in turn might have a shortcoming in a different situation. 
This shows that our method is not only robust towards unobservable degrees of cooperativeness, but also against previously unencountered policies. 

In Table~\ref{tab:mean_performance_medium_test} we also show the results of our baselines. 
While performance of the decentrally trained policies and reachability analysis is not too far apart, the negotiation capabilities vary greatly. 
We found the predominant failure mode of the baselines to be quasi point-symmetric initializations where both agents approaching from either side have the same set of behavior options available. Our learned policies are able to break this symmetry and negotiate the lane usage, even when they are controlled by the same driving policy and exhibit equivalent degrees of cooperativeness. We assume this to be due to their ability to observe details of the opposing vehicle’s movements, allowing a more refined behavior selection than the high-level features of the baselines shown in Fig.~\ref{fig:reach} do.

\setlength{\textfloatsep}{0pt}
\begin{algorithm}[t]
\SetAlgoLined
    Initialize $Q_{\theta}$ with random weights $\theta$\par
    Initialize target $\widehat{Q}_{\theta^-}$ with weights $\theta^- = \theta$\par
    Initialize policy $\pi_{\psi}$ with random weights $\psi$\par
    Initialize replay memory $D$ to capacity $N$\par
    \For{$episode = 1, M$}{
    receive initial state $s_t = s_0$\par
        \For{$t = 1, T$}{
            sample action $a_t \sim \pi$ \par
            Execute action $a_t$ in a simulator and observe reward $r_t$, state $s_{t+1}$, and observation $o_{t+1}$ \par
            Store transition $(s_t, o_t, a_t, r_t, s_{t+1}, o_{t+1})$ in $D$\par
            Sample random mini-batch of transitions $(s_i , o_i, a_i , r_i , s_{i +1}, o_{i +1})$ from $D$ \par
            Set $y_i = 
            \begin{cases}
                r_i             & \text{if term.} \\
                r_i + \gamma \sum_{a'} \pi_{\psi}(a'|o_{i+1})[Q_{\theta^-}(s_{i+1}, a') \\ \hspace{42pt} - \alpha log \pi_{\psi}(a'|o_{i+1})]         & \text{else}
            \end{cases}$ \par
            Perform a gradient descent step on $(y_i - Q_{\theta}(s_i , a_i ))^2$ with respect to the network parameters $\theta$ \par
            Every C steps set $\widehat{Q}_{\theta^-} = Q_{\theta}$ \par
            
            \For{$update=1, \pi \text{ updates per Q update}$}{
                Sample random mini-batch ($s_i$, $o_i$) from $D$ \par
                Perform a gradient descent step on $KL (\pi_{\psi}(\cdot|o_i) | \text{softmax}(Q_{\theta}(s_i, \cdot)/\alpha))$ with respect to the policy parameters $\psi$ \par
            }
        }
    }
\caption{Discrete Asymmetric Soft Actor-Critic}
\label{algoithm:dSAC}
\end{algorithm}

\section{Conclusion and Future Work}
In this paper, we introduced a new high-conflict driving scenario challenging the social component of automated cars. 
In the model of our problem, we systematically removed assumptions common in MARL to make our approach more similar to real-world driving.
We showed that, even with basic low-level control strategies, a refined behavior policy can exceed success rates of $99\%$ when negotiating bi-directional lane usage.
We showed that our parametrized reward function leads to agents exhibiting varying but interpretable behaviors. 
We found our agents to be robust to an unknown decision timing, an unobservable degree of cooperativeness of the opposing vehicle, and previously unencountered policies. 

Throughout this work, low-level controls are realized using conventional control methods, as we focus on behavior decisions. However, agents can only infer intentions of the opposing vehicle through observing the actions executed by the low-level controls. 
Learned throttle and steering control can enable a richer way of representing high-level intentions, possibly under the same reward-shaping methods as used in this work. 
In future work, devising a hierarchical RL approach might prove beneficial to decouple learning to negotiate and learning to communicate through actions at the different levels of the road user task.
For full autonomous driving, the reliable, robust, and verifiable solving of the problem presented is vital as it addresses the \textit{last-mile-problem} of the field. 

\bibliographystyle{IEEEtran}
\bibliography{bibliography}

\begin{thebibliography}{10}
\providecommand{\url}[1]{#1}
\csname url@samestyle\endcsname
\providecommand{\newblock}{\relax}
\providecommand{\bibinfo}[2]{#2}
\providecommand{\BIBentrySTDinterwordspacing}{\spaceskip=0pt\relax}
\providecommand{\BIBentryALTinterwordstretchfactor}{4}
\providecommand{\BIBentryALTinterwordspacing}{\spaceskip=\fontdimen2\font plus
\BIBentryALTinterwordstretchfactor\fontdimen3\font minus
  \fontdimen4\font\relax}
\providecommand{\BIBforeignlanguage}[2]{{%
\expandafter\ifx\csname l@#1\endcsname\relax
\typeout{** WARNING: IEEEtran.bst: No hyphenation pattern has been}%
\typeout{** loaded for the language `#1'. Using the pattern for}%
\typeout{** the default language instead.}%
\else
\language=\csname l@#1\endcsname
\fi
#2}}
\providecommand{\BIBdecl}{\relax}
\BIBdecl

\bibitem{carolyn-hierarchy}
\BIBentryALTinterwordspacing
Z.~Qiao, K.~Muelling, J.~M. Dolan, P.~Palanisamy, and P.~Mudalige, ``{POMDP}
  and hierarchical options {MDP} with continuous actions for autonomous driving
  at intersections,'' in \emph{21st International Conference on Intelligent
  Transportation Systems, {ITSC} 2018, Maui, HI, USA, November 4-7, 2018},
  W.~Zhang, A.~M. Bayen, J.~J. {S{'{a}}nchez Medina}, and M.~J. Barth,
  Eds.\hskip 1em plus 0.5em minus 0.4em\relax {IEEE}, 2018, pp. 2377--2382.
  [Online]. Available: \url{https://doi.org/10.1109/ITSC.2018.8569400}
\BIBentrySTDinterwordspacing

\bibitem{8968565}
Y.~{Chen}, C.~{Dong}, P.~{Palanisamy}, P.~{Mudalige}, K.~{Muelling}, and J.~M.
  {Dolan}, ``Attention-based hierarchical deep reinforcement learning for lane
  change behaviors in autonomous driving,'' in \emph{2019 IEEE/RSJ
  International Conference on Intelligent Robots and Systems (IROS)}, 2019, pp.
  3697--3703.

\bibitem{oliehoek_amato_2016}
F.~A. Oliehoek and C.~Amato, \emph{A Concise Introduction to Decentralized
  POMDPs}.\hskip 1em plus 0.5em minus 0.4em\relax Springer International
  Publishing, 2016.

\bibitem{webb07}
J.~N. Webb, \emph{Game Theory - Decisions, Interaction and Evolution},
  1st~ed.\hskip 1em plus 0.5em minus 0.4em\relax London: Springer, 2007.

\bibitem{10.5555/3305381.3305521}
T.~Haarnoja, H.~Tang, P.~Abbeel, and S.~Levine, ``Reinforcement learning with
  deep energy-based policies,'' in \emph{Proceedings of the 34th International
  Conference on Machine Learning - Volume 70}, 2017, p. 1352–1361.

\bibitem{mnih2015humanlevel}
V.~Mnih, K.~Kavukcuoglu, D.~Silver, A.~A. Rusu, J.~Veness, M.~G. Bellemare,
  A.~Graves, M.~Riedmiller, A.~K. Fidjeland, G.~Ostrovski, S.~Petersen,
  C.~Beattie, A.~Sadik, I.~Antonoglou, H.~King, D.~Kumaran, D.~Wierstra,
  S.~Legg, and D.~Hassabis, ``Human-level control through deep reinforcement
  learning,'' \emph{Nature}, vol. 518, no. 7540, pp. 529--533, Feb. 2015.

\bibitem{10.5555/3237383.3238080}
P.~Sunehag, G.~Lever, A.~Gruslys, W.~M. Czarnecki, V.~Zambaldi, M.~Jaderberg,
  M.~Lanctot, N.~Sonnerat, J.~Z. Leibo, K.~Tuyls, and T.~Graepel,
  ``Value-decomposition networks for cooperative multi-agent learning based on
  team reward,'' in \emph{Proceedings of the 17th International Conference on
  Autonomous Agents and MultiAgent Systems}.\hskip 1em plus 0.5em minus
  0.4em\relax Richland, SC: International Foundation for Autonomous Agents and
  Multiagent Systems, 2018, p. 2085–2087.

\bibitem{DBLP:conf/icml/Tan93}
M.~Tan, ``Multi-agent reinforcement learning: Independent versus cooperative
  agents,'' in \emph{Machine Learning, Proceedings of the Tenth International
  Conference, University of Massachusetts, Amherst, MA, USA, June 27-29, 1993},
  P.~E. Utgoff, Ed.\hskip 1em plus 0.5em minus 0.4em\relax Morgan Kaufmann,
  1993, pp. 330--337.

\bibitem{10.1371/journal.pone.0172395}
A.~Tampuu, T.~Matiisen, D.~Kodelja, I.~Kuzovkin, K.~Korjus, J.~Aru, J.~Aru, and
  R.~Vicente, ``Multiagent cooperation and competition with deep reinforcement
  learning,'' \emph{PLOS ONE}, vol.~12, no.~4, pp. 1--15, 04 2017.

\bibitem{10.5555/3305381.3305500}
J.~Foerster, N.~Nardelli, G.~Farquhar, T.~Afouras, P.~H.~S. Torr, P.~Kohli, and
  S.~Whiteson, ``Stabilising experience replay for deep multi-agent
  reinforcement learning,'' in \emph{Proceedings of the 34th International
  Conference on Machine Learning - Volume 70}, 2017, p. 1146–1155.

\bibitem{Rashid2018QMIXMV}
T.~Rashid, M.~Samvelyan, C.~Schroeder, G.~Farquhar, J.~Foerster, and
  S.~Whiteson, ``{QMIX}: Monotonic value function factorisation for deep
  multi-agent reinforcement learning,'' ser. Proceedings of Machine Learning
  Research, J.~Dy and A.~Krause, Eds., vol.~80.\hskip 1em plus 0.5em minus
  0.4em\relax Stockholmsmässan, Stockholm Sweden: PMLR, 10--15 Jul 2018, pp.
  4295--4304.

\bibitem{Foerster2018CounterfactualMP}
J.~N. Foerster, G.~Farquhar, T.~Afouras, N.~Nardelli, and S.~Whiteson,
  ``Counterfactual multi-agent policy gradients,'' in \emph{The Thirty-Second
  AAAI Conference on Artificial Intelligence}, 2018, pp. 2974--2982.

\bibitem{DBLP:conf/nips/LoweWTHAM17}
R.~Lowe, Y.~Wu, A.~Tamar, J.~Harb, P.~Abbeel, and I.~Mordatch, ``Multi-agent
  actor-critic for mixed cooperative-competitive environments,'' in
  \emph{Advances in Neural Information Processing Systems 30: Annual Conference
  on Neural Information Processing Systems 2017, 4-9 December 2017, Long Beach,
  CA, {USA}}, I.~Guyon, U.~von Luxburg, S.~Bengio, H.~M. Wallach, R.~Fergus,
  S.~V.~N. Vishwanathan, and R.~Garnett, Eds., 2017, pp. 6379--6390.

\bibitem{Li2019RobustMR}
S.~Li, Y.~Wu, X.~Cui, H.~Dong, F.~Fang, and S.~Russell, ``Robust multi-agent
  reinforcement learning via minimax deep deterministic policy gradient,''
  \emph{Proceedings of the AAAI Conference on Artificial Intelligence},
  vol.~33, pp. 4213--4220, 07 2019.

\bibitem{Yang2020CM3}
J.~Yang, A.~Nakhaei, D.~Isele, K.~Fujimura, and H.~Zha, ``{CM3:} cooperative
  multi-goal multi-stage multi-agent reinforcement learning,'' in \emph{8th
  International Conference on Learning Representations, {ICLR} 2020, Addis
  Ababa, Ethiopia}, 2020.

\bibitem{multiagentsoft}
E.~Wei, D.~Wicke, D.~Freelan, and S.~Luke, ``Multiagent soft q-lerning,'' in
  \emph{AAAI Spring Symposium Series}, 2018, pp. 351--357.

\bibitem{10.5555/551283}
R.~S. Sutton and A.~G. Barto, \emph{Introduction to Reinforcement Learning},
  2nd~ed.\hskip 1em plus 0.5em minus 0.4em\relax Cambridge, MA, USA: MIT Press,
  2018.

\bibitem{Haarnoja2018SoftAO}
T.~Haarnoja, A.~Zhou, P.~Abbeel, and S.~Levine, ``Soft actor-critic: Off-policy
  maximum entropy deep reinforcement learning with a stochastic actor,'' ser.
  Proceedings of Machine Learning Research, J.~Dy and A.~Krause, Eds.,
  vol.~80.\hskip 1em plus 0.5em minus 0.4em\relax Stockholmsmässan, Stockholm
  Sweden: PMLR, 10--15 Jul 2018, pp. 1861--1870.

\bibitem{christodoulou2019soft}
P.~Christodoulou, ``Soft actor-critic for discrete action settings,'' 2019
  (unpublished).

\bibitem{DBLP:journals/corr/abs-1903-01365}
G.~Bacchiani, D.~Molinari, and M.~Patander, ``Microscopic traffic simulation by
  cooperative multi-agent deep reinforcement learning,'' \emph{CoRR}, vol.
  abs/1903.01365, 2019.

\bibitem{Hu2019InteractionawareDM}
Y.~Hu, A.~Nakhaei, M.~Tomizuka, and K.~Fujimura, ``Interaction-aware decision
  making with adaptive strategies under merging scenarios,'' \emph{2019
  IEEE/RSJ International Conference on Intelligent Robots and Systems (IROS)},
  pp. 151--158, 2019.

\bibitem{8916951}
L.~{Schester} and L.~E. {Ortiz}, ``Longitudinal position control for highway
  on-ramp merging: A multi-agent approach to automated driving,'' in \emph{2019
  IEEE Intelligent Transportation Systems Conference (ITSC)}, 2019, pp.
  3461--3468.

\bibitem{Tang2019TowardsLM}
Y.~Tang, ``Towards learning multi-agent negotiations via self-play,''
  \emph{2019 IEEE/CVF International Conference on Computer Vision Workshop
  (ICCVW)}, pp. 2427--2435, 2019.

\bibitem{michon1985critical}
J.~A. Michon, ``A critical view of driver behavior models: what do we know,
  what should we do?'' in \emph{Human behavior and traffic safety}.\hskip 1em
  plus 0.5em minus 0.4em\relax Springer, 1985, pp. 485--524.

\bibitem{commonroad}
M.~{Althoff}, M.~{Koschi}, and S.~{Manzinger}, ``Commonroad: Composable
  benchmarks for motion planning on roads,'' in \emph{2017 IEEE Intelligent
  Vehicles Symposium (IV)}, 2017, pp. 719--726.

\bibitem{reachability}
M.~{Althoff} and J.~M. {Dolan}, ``Online verification of automated road
  vehicles using reachability analysis,'' \emph{IEEE Transactions on Robotics},
  vol.~30, no.~4, pp. 903--918, 2014.

\bibitem{4282788}
G.~M. {Hoffmann}, C.~J. {Tomlin}, M.~{Montemerlo}, and S.~{Thrun}, ``Autonomous
  automobile trajectory tracking for off-road driving: Controller design,
  experimental validation and racing,'' in \emph{2007 American Control
  Conference}, 07 2007, pp. 2296--2301.

\bibitem{DBLP:journals/corr/abs-1906-04737}
G.~Papoudakis, F.~Christianos, A.~Rahman, and S.~V. Albrecht, ``Dealing with
  non-stationarity in multi-agent deep reinforcement learning,'' \emph{CoRR},
  vol. abs/1906.04737, 2019.

\end{thebibliography}

\end{document}